\documentclass{article}
\usepackage{amsmath}
\usepackage{graphicx} 
\usepackage{subcaption}
\usepackage{listings}
\usepackage{amssymb}
\usepackage{cite}
\usepackage{natbib}
\usepackage{tabularx}
\usepackage{booktabs}
\usepackage[hyphens]{url}

\usepackage{xcolor}
\usepackage[margin=1in]{geometry} 
\title{Least-Squares-Embedded Optimization for Accelerated Convergence of PINNs in Acoustic Wavefield Simulations}
\author{}
\date{}

\author{
	\normalsize Mohammad Mahdi Abedi\textsuperscript{1},  
	David Pardo\textsuperscript{2,1,3},  
	Tariq Alkhalifah\textsuperscript{4} \\  
	\footnotesize  
	\textsuperscript{1}Basque Center for Applied Mathematics, Bilbao, Spain \\  
	\footnotesize  
	\textsuperscript{2}University of the Basque Country, Department of Mathematics, Spain \\  
	\footnotesize  
	\textsuperscript{3}Ikerbasque, Basque Foundation for Science, Bilbao, Spain \\  
	\footnotesize  
	\textsuperscript{4}King Abdullah University of Science and Technology, Thuwal 23955-6900, Saudi Arabia \\  
	\footnotesize  
	\textit{Emails:} mabedi@bcamath.org, david.pardo@ehu.es, tariq.alkhalifah@kaust.edu.sa  
}

\begin{document}
	\graphicspath{
	{"Figures/"}}
	\maketitle
\footnotetext{This is a preprint.}
\begin{abstract}
	Physics-Informed Neural Networks (PINNs) have shown promise in solving partial differential equations (PDEs), including the frequency-domain Helmholtz equation. However, standard training of PINNs using gradient descent (GD) suffers from slow convergence and instability, particularly for high-frequency wavefields. For scattered acoustic wavefield simulation based on Helmholtz equation we derive a hybrid optimization framework that accelerates training convergence by embedding a least-squares (LS) solver directly into the GD loss function. This formulation enables optimal updates for the linear output  layer. Our method is applicable with or without perfectly matched layers (PML), and we provide practical tensor-based implementations for both scenarios. Numerical experiments on benchmark velocity models demonstrate that our approach achieves faster convergence, higher accuracy, and improved stability compared to conventional PINN training. In particular, our results show that the LS-enhanced method converges rapidly even in cases where standard GD-based training fails. The LS solver operates on a small normal matrix, ensuring minimal computational overhead and making the method scalable for large-scale wavefield simulations.
\end{abstract}
\vspace{1em}
\section{Introduction}

Wavefield simulation is fundamental in geophysical exploration, with the Helmholtz equation governing frequency-domain wavefield solutions \citep{Alkhalifah2021, Cuomo2022}. Traditional numerical methods, such as finite-difference and finite-element approaches, face challenges in handling complex boundary conditions, high frequencies, and large-scale models due to computational costs and dispersion errors \citep{Wu2018, ElSayed2004}. Although frequency-domain formulations reduce dimensionality compared to time-domain methods, they require solving large linear systems (often via matrix inversions) which becomes computationally intensive for large or complex models \citep{Pratt1999, Sirgue2008}.

Physics-Informed Neural Networks (PINNs) have emerged as a promising alternative, employing automatic differentiation to enforce PDE constraints while providing continuous, mesh-free solutions \citep{Raissi2019}. They excel in irregular domains and adaptive sampling, but their training efficiency diminishes for high-frequency wavefields due to the inherent low-frequency bias of neural networks \citep{Neal2019, Huang2023}. They usually struggle to capture the oscillatory nature of wavefields without specialized architectures or loss functions \citep{Alkhalifah2021, Song2021}. Recent advances in high-frequency representation and model robustness  address these limitations  \citep{song2023,yang2023fwigan,waheed2022kronecker}.

Moreover, PINNs often suffer from slow convergence, as gradient descent (GD) updates inefficiently probe flat loss landscapes. To improve optimization efficiency, \citet{Cyr2020} proposed a hybrid optimization approach for feed-forward neural networks with a linear output layer. Their key insight was to treat the neurons of the last hidden layer as basis functions, with the output layer performing a linear combination of these functions. When the loss is quadratic with respect to the weights of the last layer, they replace conventional GD updates with a two-step alternating optimization: (1) a least-squares (LS) solver to compute output-layer weights, and (2) GD updates for the remaining network parameters. This approach drastically reduced the number of training iterations required for convergence by exploiting the least-squares structure of the output-layer subproblem. Subsequent work by \citet{uriarte2025} and \citet{baharlouei2025least} further developed these optimization methods for variational PINNs and parametric PDEs.

Building upon these foundations, we derive the formulation and practical implementation of an LS solver embedded into the optimization process of PINNs for solving the scattered Helmholtz equation. We embed the LS solver as part of the loss function evaluation, ensuring consistency between the optimization steps while eliminating the need for derivative recalculations that alternating optimization methods typically require. Our numerical tests show that this unified approach improves training efficiency and enhances convergence properties, particularly for high-frequency wavefields where standard PINN methods struggle.

The remainder of this paper is organized as follows: the Methodology section presents the mathematical formulation of the scattered Helmholtz equation, including the Perfectly Matched Layer, and introduces the proposed LS formulation. The Implementation section details practical aspects such as network architecture, derivative computation, and the LS solution strategy. The Numerical Results section evaluates the method's performance through experiments on benchmark velocity models, comparing convergence rates and accuracy against conventional PINN implementations. Finally, we discuss the implications of our findings, the limitations, and potential directions for future research.

\section{Methodology}
\subsection{Governing Equations}

This study considers the simulation of scattered acoustic wavefields by solving the scattered Helmholtz equation:  
\begin{equation}
	\left(\nabla^2 + \frac{\omega^2}{v^2(\mathbf{x})}\right) u_s(\mathbf{x}) = -\omega^2 \delta m(\mathbf{x}) u_0(\mathbf{x}),
	\label{eq:pde}
\end{equation}  
where the total wavefield \(u(\mathbf{x})\) is decomposed into a background wavefield \(u_0(\mathbf{x})\) and a scattered wavefield, defined as \(u_s(\mathbf{x}) = u(\mathbf{x}) - u_0(\mathbf{x})\). Here, \(\mathbf{x}\) denotes the spatial coordinates (e.g., in a two-dimensional domain, \(\mathbf{x} = \{x, z\}\)).  The perturbation in the squared slowness, given by \(\delta m(\mathbf{x}) = 1/v^2(\mathbf{x}) - 1/v_0^2\), represents the difference between the inverse squared velocity field and the inverse squared background velocity.  The background wavefield \(u_0(\mathbf{x})\) is computed analytically for a homogeneous medium with a constant background velocity \(v_0\), which is set to the velocity at the source location \citep{Alkhalifah2021}.  

To simulate wave propagation in an infinite medium while limiting the computational domain,  absorbing boundary conditions are used. The Perfectly Matched Layer (PML) technique introduces an artificial damping region around the simulation boundaries, gradually absorbing wave energy to prevent reflections \citep{berenger1994perfectly}.
With the PML applied, the scattered Helmholtz equation takes the following form \citep{wu2023,abedi2025gabor}:  
\begin{equation}
	\frac{\partial}{\partial x} \left( e_1(\mathbf{x}) \frac{\partial u_s(\mathbf{x})}{\partial x} \right) +
	\frac{\partial}{\partial z} \left( e_2(\mathbf{x}) \frac{\partial u_s(\mathbf{x})}{\partial z} \right) +
	e_3(\mathbf{x}) \omega^2 m(\mathbf{x}) u_s(\mathbf{x}) = - e_3(\mathbf{x}) \omega^2 \delta m(\mathbf{x}) u_0(\mathbf{x}),
	\label{eq:pde_pml}
\end{equation}  
where the complex coordinate stretching factors \( e_1 \), \( e_2 \), and \( e_3 \) are defined as:  
\begin{equation}
	\begin{aligned}
		e_1 &= \frac{1 + c^2 l_x^2 l_z^2}{1 + c^2 l_x^4} + i \frac{c (l_x^2 - l_z^2)}{1 + c^2 l_x^4}, \\
		e_2 &= \frac{1 + c^2 l_x^2 l_z^2}{1 + c^2 l_z^4} + i \frac{c ( l_z^2-l_x^2 )}{1 + c^2 l_z^4}, \\
		e_3 &= 1 - c^2 l_x^2 l_z^2 - i c( l_x^2 +  l_z^2).
	\end{aligned}
	\label{eq:stretching_factors}
\end{equation}
The boundary distance profiles are given by:  
\begin{equation}
	\begin{aligned}
		l_x &= \max(0, x_{bl} - x) + \max(0, x - x_{br}), \\
		l_z &= \max(0, z_{bu} - z) + \max(0, z - z_{bd}),
	\end{aligned}
\end{equation}
and the damping coefficient is defined as:  
\begin{equation}
	c = a_0 \frac{\omega_0}{\omega L_{\text{PML}}^2},
\end{equation}  
where $L_{\text{PML}}$ is the thickness of the absorbing layer, $a_0$ is a scaling constant, and $x_{bl}, x_{br}, z_{bu}, z_{bd}$ denote the positions of the left, right, upper, and lower boundaries, respectively. These stretching factors modify the governing equation to absorb outgoing waves.

The background wavefield $u_0(\mathbf{x})$ in Eq.~\eqref{eq:pde_pml} should satisfy the Helmholtz equation with PML included. An analytical approximation for the background wavefield is given as \citep{abedi2025gabor}:  
\begin{equation}
	u_0(\mathbf{x}) = \frac{i}{4} H_0^{(2)}\left(\frac{\omega}{v_0}  |\mathbf{x} - \mathbf{x}_s|\right) e^{-\omega c \frac{(l_x^2+l_z^2)}{3 v_0}^{(3/2)}},
\end{equation}  
where $H_0^{(2)}(.)$ represents the zero-order Hankel function of the second kind. This formulation is exact within the domain of interest and serves as an approximation in the PML region.

\subsection{Simple PINN}

\begin{figure}[]
	\centering
		\includegraphics[width=0.5\textwidth]{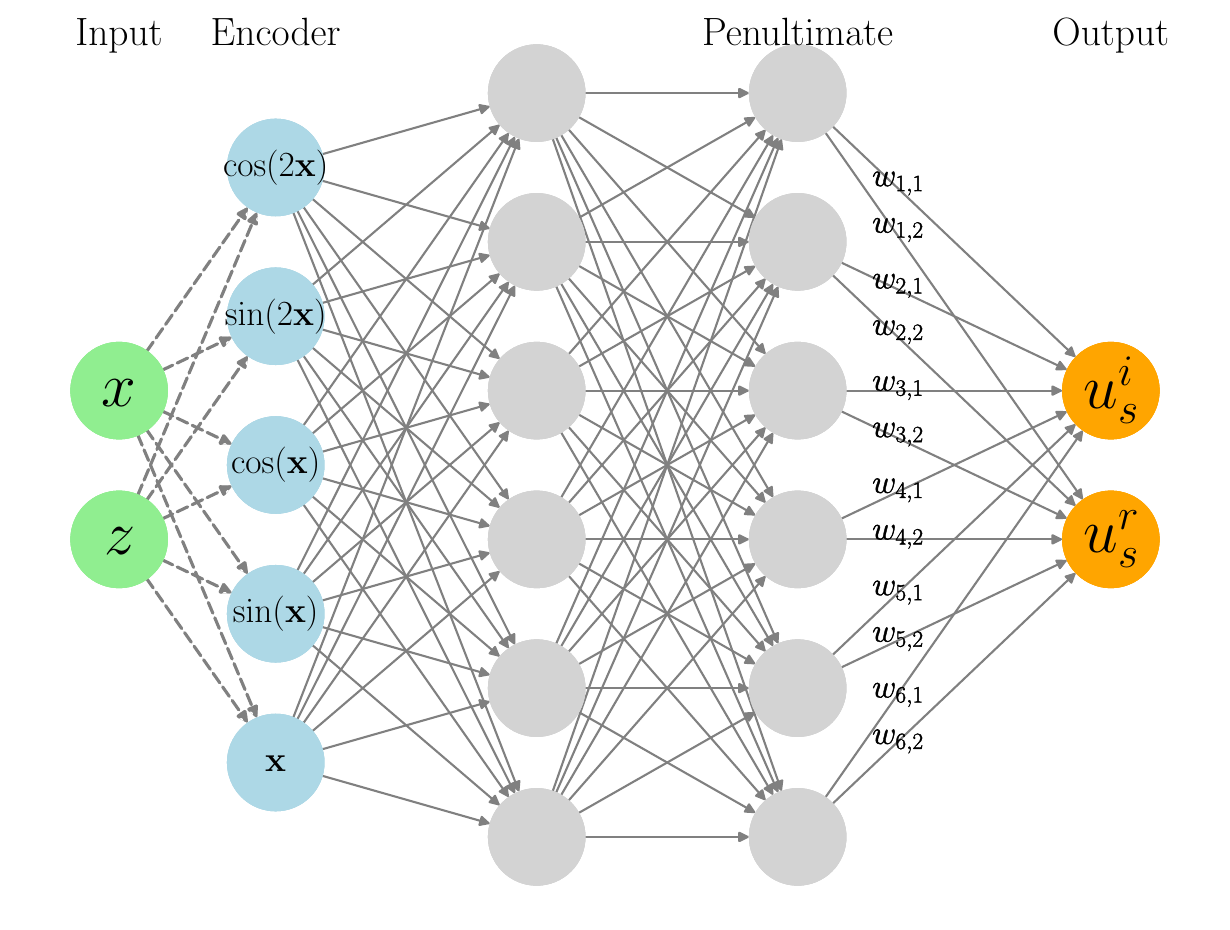}  %
	\caption{Schematic representations of the neural network architectures. \(\mathit{w}_{i,j}\) show the elements of the output layer's weights matrix \(\mathbf{W}\),  calculated by the LS solver.  Dashed connections have no associated weight.}
	\label{fig:NN_architecture}
\end{figure}

Our physics-informed neural network (PINN) is a fully connected feedforward architecture that maps the input coordinates \(\mathbf{x} = \{x, z\}\) to the real and imaginary parts of the scattered wavefield, \(u_s^r\) and \(u_s^i\) (Figure~\ref{fig:NN_architecture}). The transformation at layer \(m\) is defined as:
\begin{equation}
	\mathbf{h}^{(m)} = \phi \left( \mathbf{W}^{(m)} \mathbf{h}^{(m-1)} + \mathbf{b}^{(m)} \right),
\end{equation}
where \(\mathbf{W}^{(m)}\) and \(\mathbf{b}^{(m)}\) are the trainable weights and biases of layer \(m\), and \(\phi(\cdot)\) is an activation function. We use the sine function (\(\phi = \sin\)) in all hidden layers. The output layer is linear and does not apply any activation.

The network parameters are updated by minimizing a loss function \(\mathcal{L}\) using gradient descent (GD). The GD update rule for the weights is:
\begin{equation}
	\mathbf{W}^{(m)} \leftarrow \mathbf{W}^{(m)} - \eta \nabla_{\mathbf{W}^{(m)}} \mathcal{L},
\end{equation}
where \(\eta\) is the learning rate and \(\nabla_{\mathbf{W}^{(m)}} \mathcal{L}\) is the gradient of the loss with respect to the weights. The biases \(\mathbf{b}^{(m)}\) are updated similarly.

To improve the network's ability to capture oscillatory wavefields, we apply a sinusoidal positional encoding to the spatial coordinates \( (x, z) \). This encoding projects the input into a higher-dimensional space, allowing the network to represent high-frequency variations more effectively \citep{Vaswani2017, huang2021}. The encoded input is given by:

\begin{equation}
	\mathbf{E}(\mathbf{x}) = \left[x, z, \sin(2^k x), \cos(2^k x), \sin(2^k z), \cos(2^k z) \right]_{k=0}^{K},
	\label{eq:encoder}
\end{equation}
where \( \mathbf{E}(\mathbf{x}) \in \mathbb{R}^{2(K+1)} \), and \( K \) is the maximum frequency considered in the encoding. This transformation introduces a multi-scale representation of the input. A schematic of the encoder layer is shown in Figure \ref{fig:NN_architecture}.

\subsection{Loss Function}
The loss function enforces the governing PDE while incorporating a soft constraint to enhance solution stability. The total loss \( \mathcal{L} \) consists of the PDE residual loss (\(\mathcal{L}_{\text{PDE}}\)) plus a constraint term (\(\mathcal{L}_{\text{C}}\)) weighted by a constant factor \(\beta\):  
\begin{equation}
	\mathcal{L} = \mathcal{L}_{\text{PDE}} + \beta \mathcal{L}_{\text{C}}.
	\label{eq:loss_total}
\end{equation}

\paragraph{PDE Loss Without PML:} 
In the absence of a perfectly matched layer (PML) and for a real-valued (non-attenuative) velocity model, the PDE loss naturally separates into two terms, enforcing the real and imaginary parts of  the scattered wavefield to satisfy the Helmholtz equation \citep{Alkhalifah2021}:
\begin{equation}
	\begin{aligned}
		\mathcal{L}_{\text{PDE}} = \frac{1}{N} \sum_{j=1}^{N} \left( 
		\left| \nabla^2 u_s^r(\mathbf{x}_j) + \frac{\omega^2}{v^2(\mathbf{x}_j)} u_s^r(\mathbf{x}_j) + \omega^2 \delta m(\mathbf{x}_j) u_{0}^r(\mathbf{x}_j) \right|^2 
		\right. + \\
		\left. \left| \nabla^2 u_s^i(\mathbf{x}_j) + \frac{\omega^2}{v^2(\mathbf{x}_j)} u_s^i(\mathbf{x}_j) + \omega^2 \delta m(\mathbf{x}_j) u_{0}^i(\mathbf{x}_j) \right|^2 \right),
	\end{aligned}
	\label{eq:loss_pde}
\end{equation}
where \( N \) is the number of collocation points. Both terms share the background wavefield \( u_0 \), which links them.

\paragraph{PDE Loss With PML:}  
With the PML incorporated, the scattered wavefield \( u_s(\mathbf{x}_j) \) must satisfy the modified governing equation, which introduces complex stretching factors in the coordinate system. The loss function \(\mathcal{L}_{\text{PDE}}\) is defined as the mean squared error of the modified PDE residual over \( N \) collocation points \citep{abedi2025gabor}:  

\begin{equation}
	\mathcal{L}_{\text{PDE}} = \frac{1}{N} \sum_{j=1}^{N} \left|  
	\frac{\partial}{\partial x} \left( e_1 \frac{\partial u_s}{\partial x} \right) +
	\frac{\partial}{\partial z} \left( e_2 \frac{\partial u_s}{\partial z} \right) +
	e_3 \omega^2 \left( m(\mathbf{x}_j) u_s(\mathbf{x}_j) + \delta m(\mathbf{x}_j) u_0(\mathbf{x}_j) \right)
	\right|^2.
	\label{eq:loss_pde_pml}
\end{equation}  
Unlike the case without PML, this formulation couples the real and imaginary parts of the wavefield, preventing their separation into two terms.  This formulation forces the network to satisfy the scattered wavefield \( u_s(\mathbf{x}) \) while incorporating the effects of the PML to attenuate the outgoing waves.

\paragraph{The soft constraint term:} 
\(\mathcal{L}_{C}\) minimizes the scattered wavefield near the source to prevent the trivial solution \(u_s = -u_0\) with a large magnitude close to the source \citep{Huang2023}. It is defined as:
\begin{equation}
	\mathcal{L}_{C} = \frac{1}{N_{C}} \sum_{j=1}^{N_{C}} \left| u_s(\mathbf{x}_{C,j}) \cdot \gamma(\mathbf{x}_{C,j})\right|^2,
\end{equation}
where \(\mathbf{x}_{C,j}\) are the \(N_{C}\) collocation points within the constraint radius, and \(\gamma(\mathbf{x}_{C,j})\) is a scaling factor:
\begin{equation}
	\gamma(\mathbf{x}_{C,j}) =\sqrt{ \max\left(0, \left({\lambda}/{4}\right) ^2- r_f^2 \right)},
	\label{eq:gama}
\end{equation}
where \(r_f\) is the distance from the source. This factor imposes a stronger penalty near the source, decreasing as \(r_f\) approaches \(\lambda/4\).

\subsection{Least Squares Solver for the Output Layer}

When the output layer is linear (i.e., without an activation function) and the loss is quadratic, its optimal weights (in the current state of the network) can be computed using a least squares (LS) solver \citep{Cyr2020}. Unlike gradient descent (GD), which iteratively updates weights based on loss gradients, the LS solver directly minimizes the loss for the output layer in a single step by solving a linear system, effectively incorporating second-order optimization.

To implement the LS solver, we first express the output linear layer's operation (without bias) in matrix form:
\begin{equation}
	\mathbf{U}_s = \mathbf{H} \mathbf{W}.
	\label{eq:layer_matrix}
\end{equation}
This compact representation is essential for an efficient formulation of the inverse problem. Here, \( \mathbf{H} \in \mathbb{R}^{N \times P} \) is the matrix representation of $\mathbf{h}$, the outputs of the penultimate layer, and \( \mathbf{U}_s \in \mathbb{R}^{N \times 2} \) is the matrix representation of $u_s$ for a batch of collocation points.  \(N\) is the batch size and \(P\) is the penultimate layer size.  \( \mathbf{W} \) is the matrix of trainable weights of the final (output) layer.

To solve for \( \mathbf{W} \) in an LS system, we substitute this matrix representation of $u_s$ (equation \ref{eq:layer_matrix}) into the loss function (equation \ref{eq:loss_total}) to form:  
\begin{equation}
	\mathcal{L} = \sum_{j} \left| \mathbf{D} \mathbf{W} - \mathbf{R} \right|^2.
	\label{eq:LS_loss1}
\end{equation}  
where \( \mathbf{D} \) is derived from the left-hand side of the PDE (incorporating derivatives of \( \mathbf{H} \)) and the soft constraint, and \( \mathbf{R} \) represents the right-hand side, which includes the background wavefield.

\subsubsection{Formulation for the Scattered Helmholtz PDE Without PML}

For equation \ref{eq:pde}, which effectively governs the real and imaginary parts separately, we obtain:
\begin{align}
	\mathbf{D}_{\text{PDE}} &= \sqrt{\frac{1}{N}} \left( \frac{\omega^2}{v^2} \mathbf{H} + \nabla^2 \mathbf{H} \right), \quad \text{(dimensions: \( N \times P \))}, \\
	\mathbf{R}_{\text{PDE}} &= \sqrt{\frac{1}{N}} \left( -\omega^2 \delta m \mathbf{U}_0 \right), \quad \text{(dimensions: \( N \times 2 \))},
\end{align}
where \( \mathbf{U}_0 \) represents the matrix of real and imaginary parts of the background wavefield (dimensions: \( N \times 2 \)), and \( \nabla^2 \mathbf{H} \) is the Laplacian of the penultimate layer output with respect to \( \mathbf{x} \). 

Next, to incorporate the soft constraint term \( \mathcal{L}_{\text{C}} \), we define:
\begin{equation}
	\mathbf{D}_{\text{C}} = \sqrt{\frac{\beta}{N_{\text{C}}}} \mathbf{H}_{\text{C}} \circ \Gamma, \quad \text{(dimensions: \( N_{\text{C}} \times P \))},
\end{equation}
where \( \mathbf{H}_{\text{C}} \) is the output of the penultimate layer for \( N_{\text{C}} \) collocation points near the source, and \( \Gamma \) is a matrix formed by repeating the \( \gamma \)-vectors (defined in equation \ref{eq:gama}) \( P \) times. The symbol \( \circ \) denotes elementwise multiplication.
The final definitions of \( \mathbf{D} \) and \( \mathbf{R} \), read:
\begin{equation}
	\mathbf{D} = \begin{bmatrix}
		\mathbf{D}_{\text{PDE}} \\
		\mathbf{D}_{\text{C}}
	\end{bmatrix}, \quad 
	\mathbf{R} = \begin{bmatrix}
		\mathbf{R}_{\text{PDE}} \\
		\mathbf{0}
	\end{bmatrix}.
	\label{eq:DR}
\end{equation}
The matrix \( \mathbf{D} \) has dimensions \( (N + N_{\text{C}}) \times P \), and  \( \mathbf{R} \) has dimensions \( (N + N_{\text{C}}) \times 2 \). 

The optimal weights \( \mathbf{W}^* \) are obtained using a damped LS solution:  
\begin{equation}
	\mathbf{W}^* = \left( \mathbf{D}^T \mathbf{D} + \epsilon \mathbf{I} \right)^{-1} \mathbf{D}^T \mathbf{R},
	\label{eq:LS_solution}
\end{equation}  
where \( \epsilon \) is a small positive number that imposes a damping for numerical stability, and \( \mathbf{I} \) is the identity matrix.  \( \mathbf{W}^* \) ensures an optimal mapping from the penultimate layer’s output to the desired wavefield.

\paragraph{LS Solver as Part of the GD Loss:}  
To embed the LS solver within the itterative GD-based training, we reformulate the loss function in equation \ref{eq:loss_total}. Using the derivatives matrix \( \mathbf{D} \), we define the GD loss as the squared error between the LS-predicted left-hand side \( \mathbf{D} \mathbf{W}^* \) and the desired right-hand side \( \mathbf{R} \):  

\begin{equation}
	\mathcal{L} = \sum_{j} \left| \mathbf{D} \mathbf{W^*} - \mathbf{R} \right|^2.
	\label{eq:LS_loss}
\end{equation}  
This formulation ensures consistency between the LS solution and GD updates. The GD is calculated using the derivatives of \( \mathbf{H} \) calculated for the LS subproblem, without a recalculation of derivatives of \( u_s \) that are needed to calculate equations \ref{eq:loss_pde} or \ref{eq:loss_pde_pml}.

\subsubsection{Formulation for the Scattered Helmholtz PDE With PML}

In Equation \ref{eq:pde_pml}, the real and imaginary parts of the scattered wavefield are explicitly coupled, meaning that both \( u_s^r \) and \( u_s^i \) depend on the real and imaginary parts of the background wavefield \( u_0^r \) and \( u_0^i \). This coupling introduces additional complexity in the formulation, which, when represented in matrix form, leads to a more complex matrix with larger dimensions compared to the previous case. For the left-hand side of the equation, which involves the differential operator applied to the scattered wavefield, we obtain:

\begin{equation}
	\begin{aligned}
		\mathbf{D}_{\text{PDE}}^r &= \sqrt{\frac{1}{N}} \left( \frac{\partial e_1^r}{\partial x} \frac{\partial \mathbf{H}}{\partial x} 
		+ e_1^r \frac{\partial^2 \mathbf{H}}{\partial x^2}
		+ \frac{\partial e_2^r}{\partial z} \frac{\partial \mathbf{H}}{\partial z} 
		+ e_2^r \frac{\partial^2 \mathbf{H}}{\partial z^2}
		+ e_3^r \omega^2  m \mathbf{H}\right), \quad \text{(dimensions: \( N \times P \))}, \\
		\mathbf{D}_{\text{PDE}}^i &=  \sqrt{\frac{1}{N}} \left(- \frac{\partial e_1^i}{\partial x} \frac{\partial \mathbf{H}}{\partial x}
		- e_1^i \frac{\partial^2 \mathbf{H}}{\partial x^2}
		- \frac{\partial e_2^i}{\partial z} \frac{\partial \mathbf{H}}{\partial z}
		- e_2^i \frac{\partial^2 \mathbf{H}}{\partial z^2}
		- e_3^i \omega^2  m \mathbf{H}\right), \quad \text{(dimensions: \( N \times P \))}.
	\end{aligned}
\end{equation}
The complex coordinate stretching factors \( e_1 \), \( e_2 \), and \( e_3 \) are defined in Equation \ref{eq:stretching_factors}, with superscripts \( r \) and \( i \) indicating the real and imaginary components. For the right-hand side, which includes the background wavefield, we obtain:

\begin{equation}
	\begin{aligned}
		\mathbf{R}_1 &= \sqrt{\frac{1}{N}} \left(
		- e_3^r \omega^2  \delta m \mathbf{U}_0^r+ e_3^i \omega^2  \delta m \mathbf{U}_0^i\right), \quad \text{(dimensions: \( N \times 1 \))}, \\
		\mathbf{R}_2 &= \sqrt{\frac{1}{N}} \left(
		- e_3^r \omega^2  \delta m \mathbf{U}_0^i - e_3^i \omega^2  \delta m \mathbf{U}_0^r\right), \quad \text{(dimensions: \( N \times 1 \))}.
	\end{aligned}
\end{equation}
Finally, the system of equations can be expressed in the required matrix form:
\begin{equation}
	\mathbf{D} = \begin{bmatrix}
		\mathbf{D}_{\text{PDE}}^r & \mathbf{D}_{\text{PDE}}^i\\
		-\mathbf{D}_{\text{PDE}}^i & \mathbf{D}_{\text{PDE}}^r\\
		\mathbf{D}_{\text{C}}  & \mathbf{D}_{\text{C}}
	\end{bmatrix}, \quad 
	\mathbf{R} = \begin{bmatrix}
		\mathbf{R}_{\text{1}} \\
		\mathbf{R}_{\text{2}} \\
		\mathbf{0}
	\end{bmatrix}.
	\label{eq:DR_PML}
\end{equation}
The matrix \( \mathbf{D} \) has dimensions \( (2N + N_{\text{C}}) \times 2P \), and the right-hand side vector \( \mathbf{R} \) has dimensions \( (2N + N_{\text{C}}) \times 1 \). When substituted into Equations \ref{eq:LS_solution} and \ref{eq:LS_loss}, they result in a least-squares problem that minimizes the residual of the scattered Helmholtz PDE, incorporating PML and the soft constraint condition.

\section{Implementation}

In our implementation, the neural network (NN) is designed to output \( \mathbf{H} \) from the penultimate layer rather than the scattered wavefield \( \mathbf{U_s} \). The output \( \mathbf{H} \) is then used to construct the matrix \( \mathbf{D} \), which requires computing the derivatives of each element of \( \mathbf{H} \) with respect to the input spatial coordinates \( (x, z) \). The formulation of  \( \mathbf{D} \) is explicitly defined in the previous section. Next, we solve for the optimal output layer's weights, and calculate the gradient descent loss in each training step.  The following practical considerations are crucial for efficient and effective implementation:

\paragraph{Derivative Computation:}
To compute the derivatives of \( \mathbf{H} \) with respect to \( \mathbf{x} \), we employ forward-mode differentiation. This technique calculates derivatives via Jacobian-vector products, propagating directional derivatives through the network instead of relying on backpropagation. Unlike backpropagation, which scales with the number of output neurons from which derivatives are computed, forward-mode differentiation scales linearly with the number of input neurons \citep{uriarte2025}. As a result, forward-mode differentiation is particularly advantageous in our implementation, where the number of neurons in the penultimate layer (\( P \)) is larger than the number of input neurons.

\paragraph{Solving the Least-Squares Problem:}  
To efficiently solve the least-squares system  (Equation~\ref{eq:LS_solution}), we employ Cholesky decomposition, a numerical method designed for symmetric positive-definite systems. This approach is computationally efficient, offering a lower computational cost compared to other decomposition methods.  Cholesky decomposition remains stable when the matrix \( \mathbf{D} \) is numerically full-rank and well-conditioned. During the initial training epochs, when \( \mathbf{H} \) is generated by randomly initialized weights in the hidden layers, numerical instability may occur in the LS solver. To prevent this, we initially use a relatively large value for the Tikhonov regularization \( \epsilon \) (e.g., 0.1) in equation \ref{eq:LS_solution} to enhance numerical stability, and as the trianing progress and \( \mathbf{H} \) becomes more structured, we gradualy reduce \( \epsilon \) to a smaller value (e.g., \( 10^{-4} \)) to maintain accuracy without excessive regularization.

\paragraph{Gradient Descent:} Using automatic differentiation, the gradients of the loss with respect to the network weights are computed, and the weights are updated using an optimizer such as Adam. Since the proposed loss function in Equation~\ref{eq:LS_loss} incorporates the LS system, it is essential to ensure that gradient descent backpropagation correctly accounts for the gradients of the LS problem at each training step.

\section{Numerical Results}

We evaluate the proposed method using two benchmark velocity models, which are commonly employed in PINN-based acoustic wavefield simulations. Experiments are performed at two different frequencies to compare the convergence and accuracy between the conventional gradient descent (GD) and the proposed LS-GD training methods, both applied to a standard PINN that incorporates positional encoding.

\begin{figure}[t]
	\centering
	\begin{subfigure}{0.32\textwidth}
		\centering
		\includegraphics[width=\textwidth]{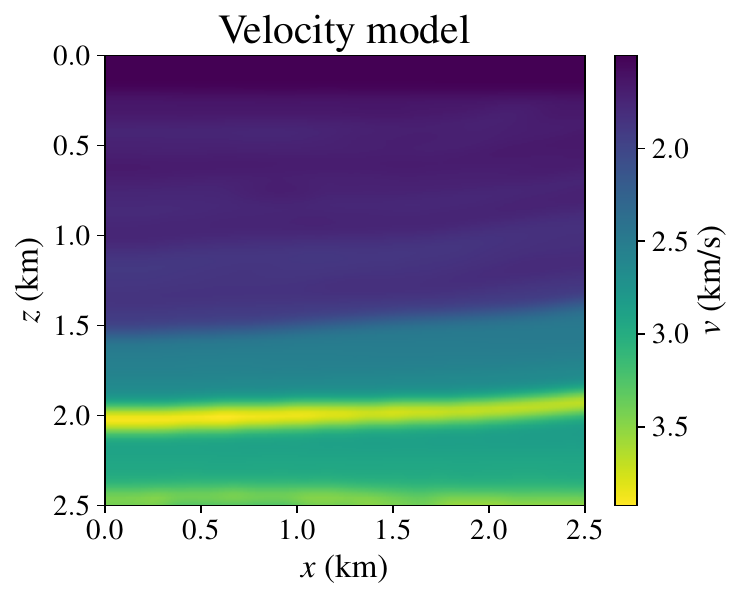}
		\caption{}
	\end{subfigure}
	\begin{subfigure}{0.32\textwidth}
		\centering
		\includegraphics[width=\textwidth]{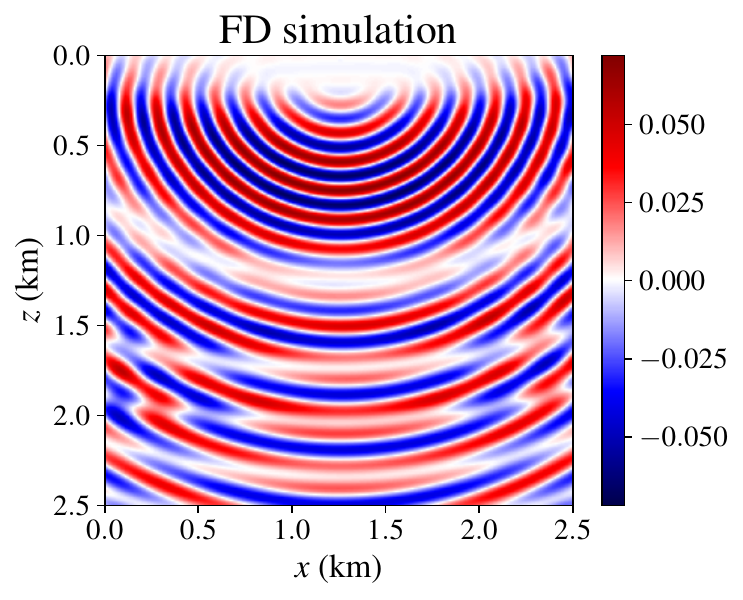}
		\caption{}
	\end{subfigure}
	
	\begin{subfigure}{0.32\textwidth}
		\centering
		\includegraphics[width=\textwidth]{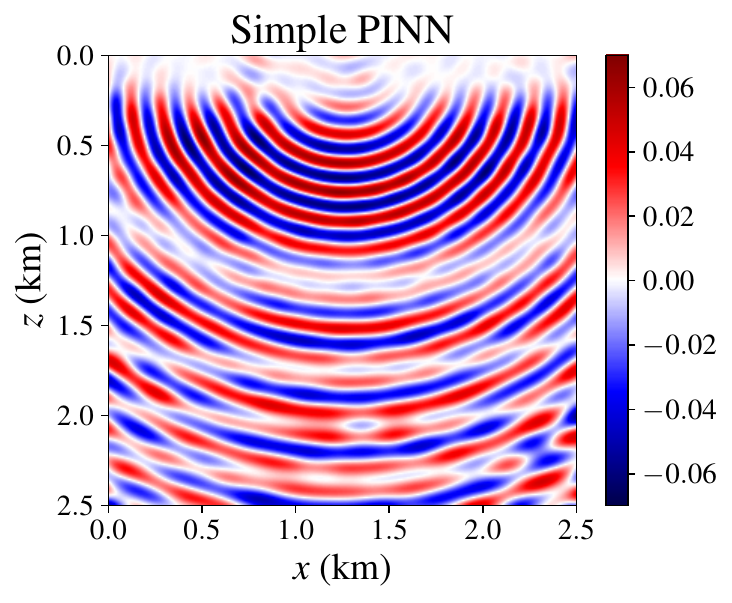}
		\caption{}
	\end{subfigure}
	\begin{subfigure}{0.32\textwidth}
		\centering
		\includegraphics[width=\textwidth]{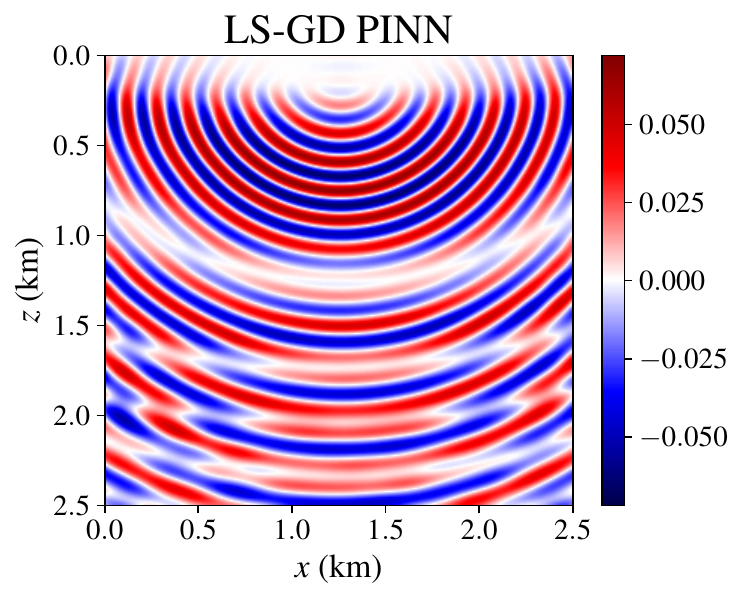}
		\caption{}
	\end{subfigure}
	
	\caption{Comparison of a 10Hz scattered wavefield predictions. (a) The velocity model used for simulation. (b) The real part of the finite-difference computed scattered wavefield. (c) Best prediction  of a simple PINN within 50,000 epochs. (d) Prediction using the proposed LS-GD PINN, which incorporates the LS solver in the training process while keeping all other parameters the same as the simple PINN.}
	\label{fig:comparison}
\end{figure}

\subsection{Simple Velocity Model}

The first test considers a simple velocity model, shown in Figure~\ref{fig:comparison}a. We simulate a 10 Hz scattered wavefield using finite difference (FD) modeling, with the source positioned at a depth of 26 meters at the model's horizontal center. The FD solution, presented on a 200\,$\times$\,200 grid, serves as the reference for PINN comparisons. To ensure accuracy, FD modeling is performed on a refined grid with four times the spatial resolution and a large perfectly matched layer (PML).

We implement a deep neural network (DNN) with four hidden layers of 64 neurons each, incorporating the encoder layer defined in Equation~\ref{eq:encoder} with $K=3$. The network is trained for 50,000 epochs using an exponentially decaying learning rate, starting at 0.002 and decreasing to approximately 0.0007. We oberved that learning rate decay improves training stability and has been shown to enhance PINN performance~\citep{bihlo2024improving, xu2024preprocessing}. 

\begin{figure}[t]
	\centering
	\begin{subfigure}{0.32\textwidth}
		\centering
		\includegraphics[width=\textwidth]{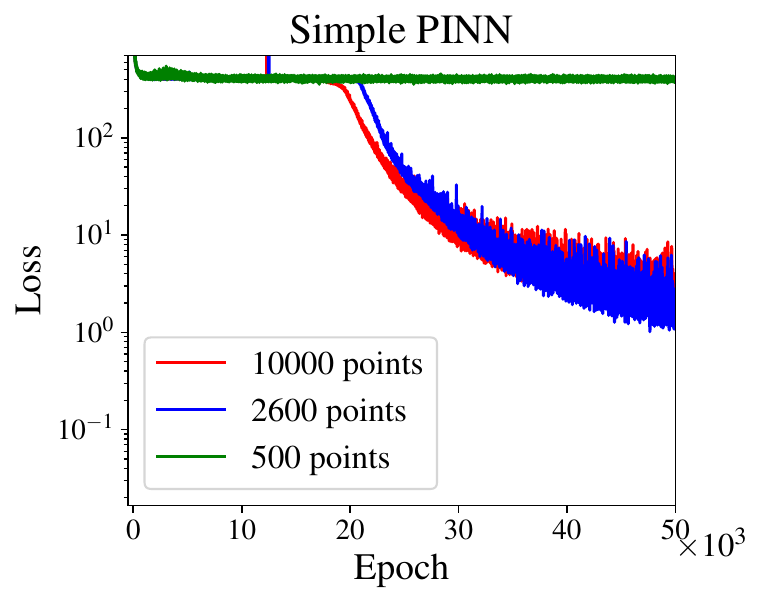}
		\caption{}
	\end{subfigure}
	\hfill
	\begin{subfigure}{0.32\textwidth}
		\centering
		\includegraphics[width=\textwidth]{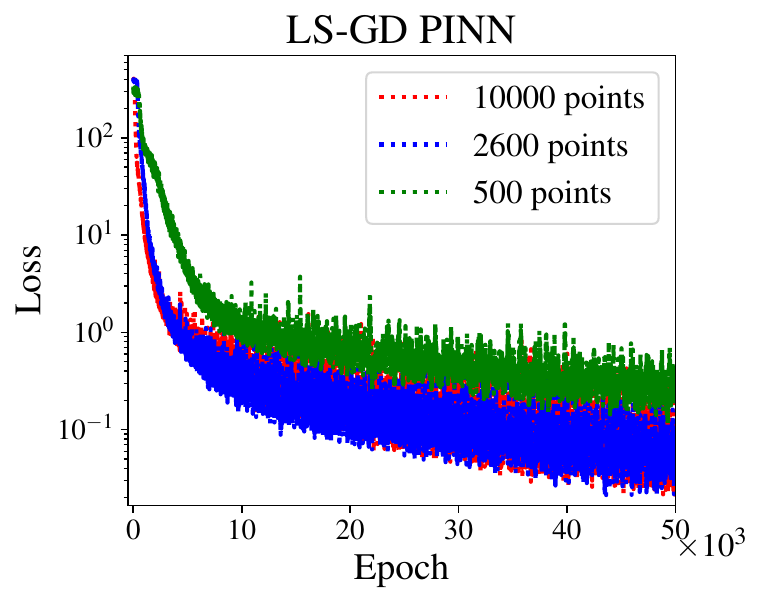}
		\caption{}
	\end{subfigure}
	\hfill
	\begin{subfigure}{0.32\textwidth}
		\centering
		\includegraphics[width=\textwidth]{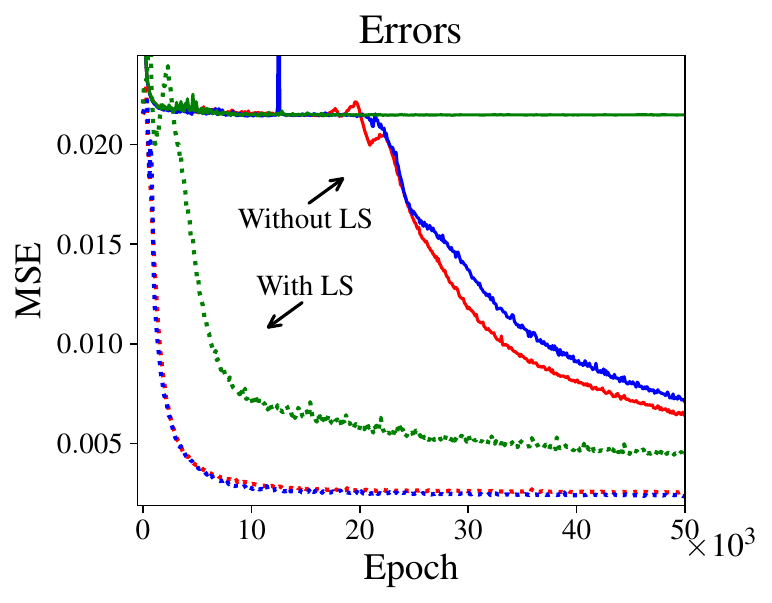}
		\caption{}
	\end{subfigure}
	
	\caption{Evolution of training losses and validation errors for different training configurations (related to Figure~\ref{fig:comparison}). Loss evolution for (a) the simple PINN, and (b) the proposed LS-GD PINN. (c) Validation errors compared to the finite difference reference solution. Training was repeated with three different collocation point counts per epoch. The LS solver improved convergence, even with very few points, where the simple PINN failed after 50,000 epochs.}
	\label{fig:ncollocation_loss_comparison}
\end{figure}

We compare two optimization methods: the simple PINN employs GD, while the proposed LS-GD PINN integrates the LS solver during training. All other parameters remain identical. Both PINN methods are trained without PML, using Equations~\ref{eq:loss_total} and \ref{eq:loss_pde} to define the loss function for the standard PINN, and Equations~\ref{eq:LS_loss} and \ref{eq:DR} for the LS-GD method. The best predictions from both methods are shown in Figure~\ref{fig:comparison}.
\paragraph{Effect of the Number of Collocation Points:}
To analyze the effect of $N$ on convergence, we train the network with 500, 2600, and 10,000 randomly selected collocation points. As shown in Figure~\ref{fig:ncollocation_loss_comparison}, the standard PINN fails to converge with 500 points and only starts converging after approximately 20,000 epochs when more collocation points are used. In contrast, the LS-GD training method enables rapid convergence from the first epochs and achieves higher final accuracy than simple gradient descent training, even when the number of collocation points is small.

\begin{figure}[t]
	\centering
	\begin{subfigure}{0.32\textwidth}
		\centering
		\includegraphics[width=\textwidth]{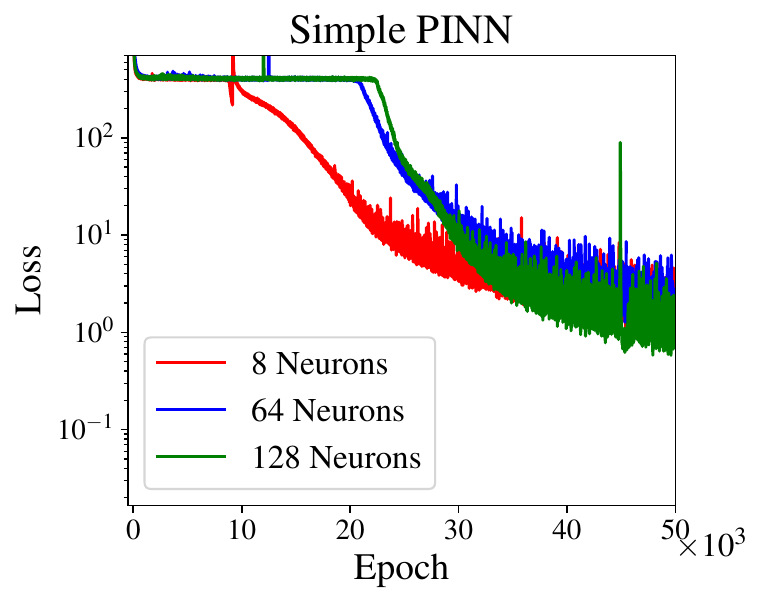}
		\caption{}
	\end{subfigure}
	\hfill
	\begin{subfigure}{0.32\textwidth}
		\centering
		\includegraphics[width=\textwidth]{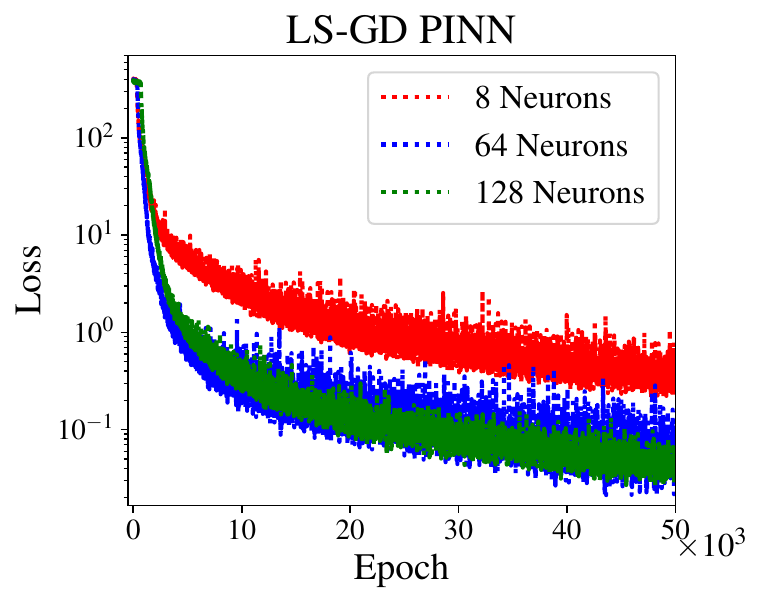}
		\caption{}
	\end{subfigure}
	\hfill
	\begin{subfigure}{0.32\textwidth}
		\centering
		\includegraphics[width=\textwidth]{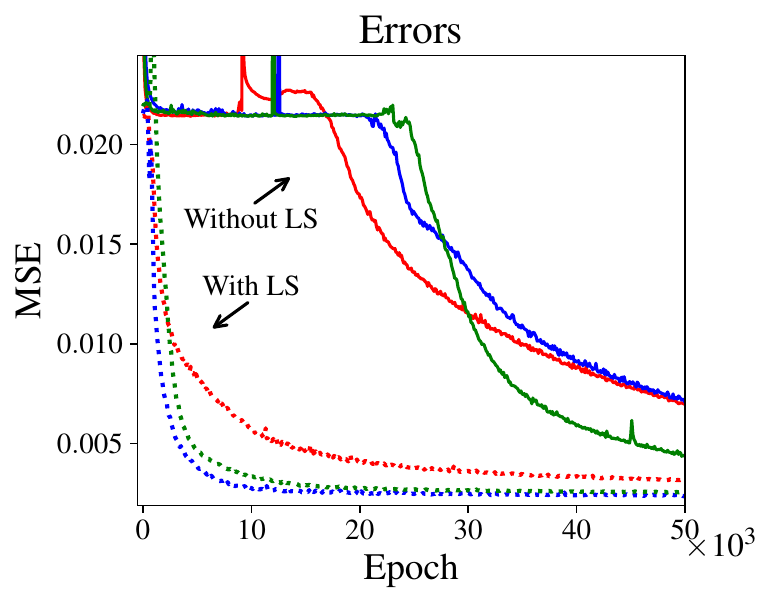}
		\caption{}
	\end{subfigure}
	\caption{Evolution of training losses and validation errors for different values of $P$, the numbers of neurons in the penultimate layer (related to Figure~\ref{fig:comparison}).  $P$ directly determines the number of weights estimated by the LS step. In all cases, the LS solver improves convergence and achieves lower errors than the standard PINN, even for smaller values of $P$.}
	\label{fig:penultimate_loss}
	
\end{figure}

\paragraph{Effect of the Penultimate Layer Size:}
In a second experiment, we analyze the impact of the number of neurons in the penultimate layer ($P$) on training loss and validation error. We compare networks with 8, 64, and 128 neurons in this layer. Since the LS solver estimates weights that are twice the number of neurons in the penultimate layer, $P$ directly influences the role of the LS solver in optimization. Figure~\ref{fig:penultimate_loss} illustrates the loss and validation error evolution for each configuration. In all cases, the inclusion of the LS solver significantly improves convergence, even when the penultimate layer contains a small number of neurons.

\subsection{Marmousi Velocity Model}

\begin{figure}[htp]
	\centering
	\begin{subfigure}{0.42\textwidth}
		\centering
		\includegraphics[width=\textwidth]{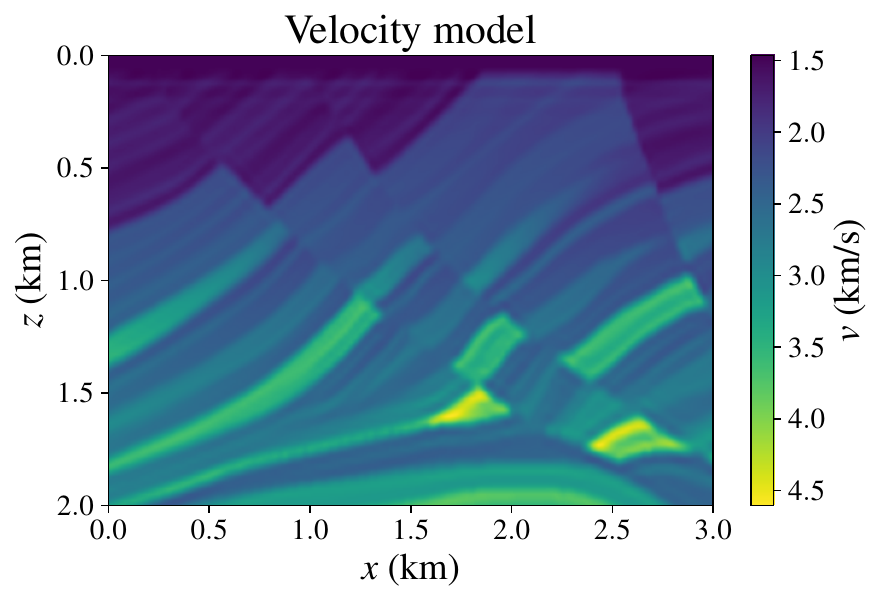}
		\label{fig:marmousi_velocity}
	\end{subfigure}
	\begin{subfigure}{0.42\textwidth}
		\centering
		\includegraphics[width=\textwidth]{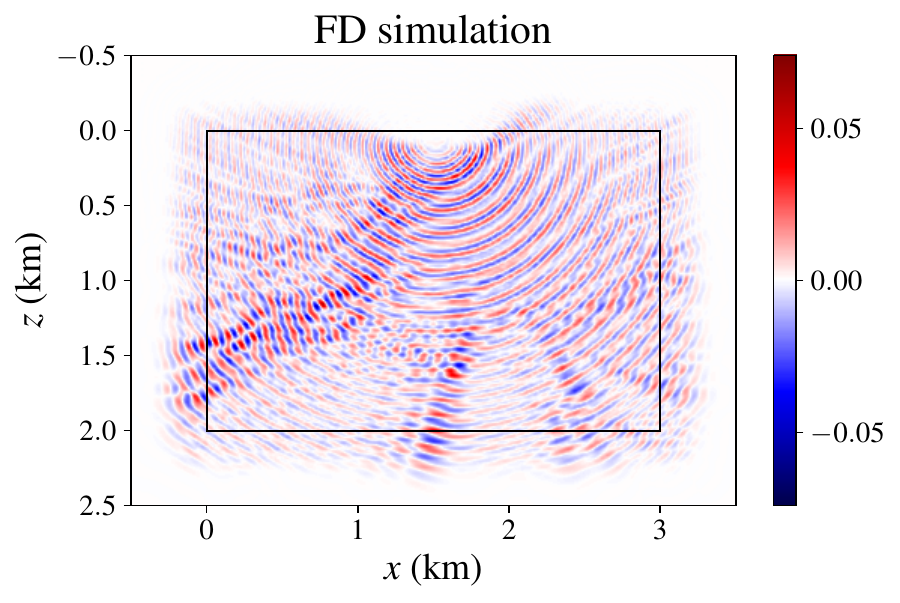}
		\label{fig:fd_solution}
	\end{subfigure}
	\begin{subfigure}{0.42\textwidth}
		\centering
		\includegraphics[width=\textwidth]{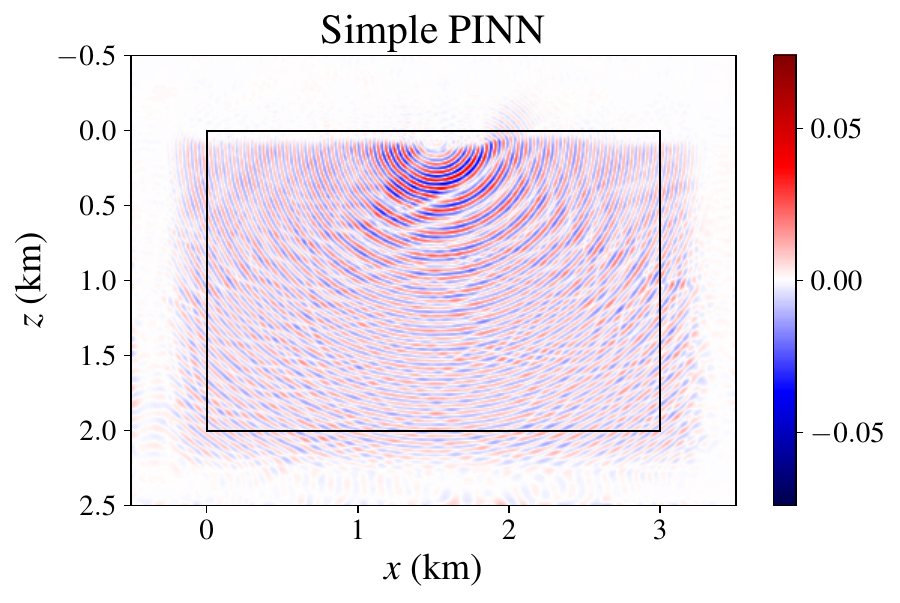}
		\label{fig:pinn_prediction}
	\end{subfigure}
	\begin{subfigure}{0.42\textwidth}
		\centering
		\includegraphics[width=\textwidth]{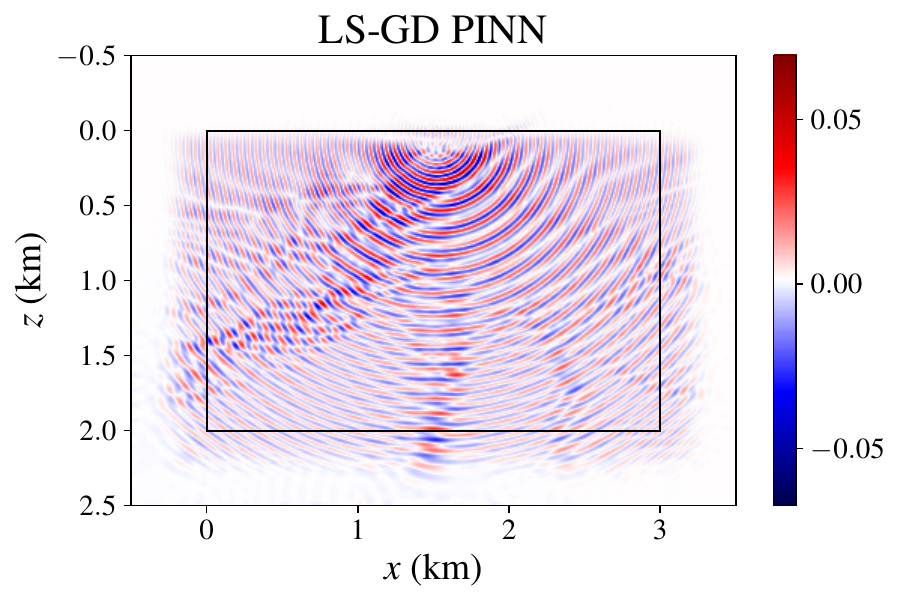}
		\label{fig:ls_gd_pinn_prediction}
	\end{subfigure}
	\begin{subfigure}{0.42\textwidth}
		\centering
		\includegraphics[width=\textwidth]{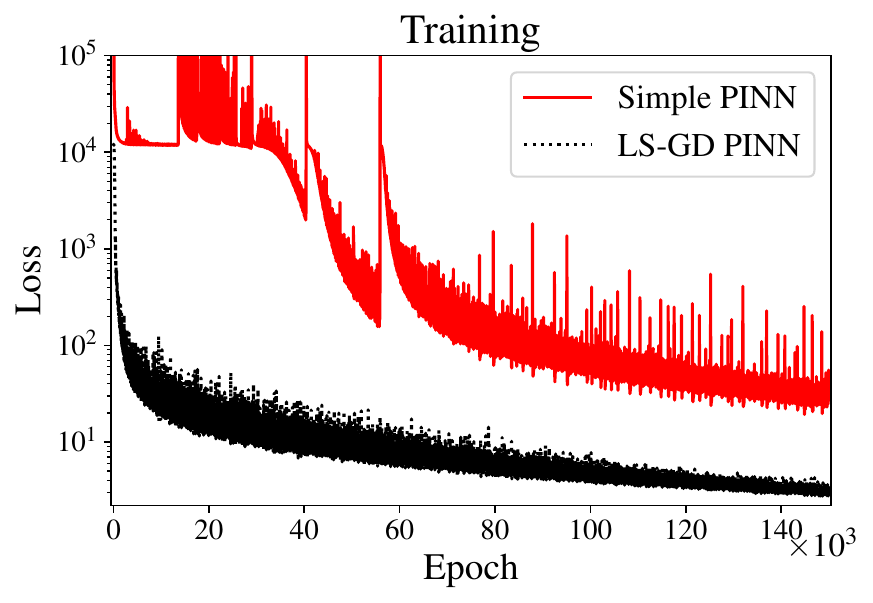}
		\label{fig:training_loss}
	\end{subfigure}
	\begin{subfigure}{0.42\textwidth}
		\centering
		\includegraphics[width=\textwidth]{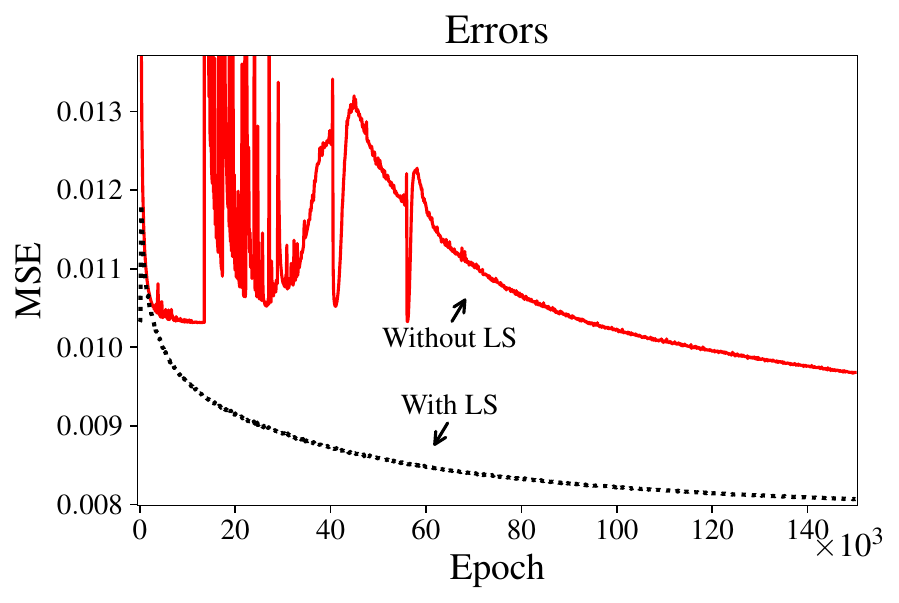}
		\label{fig:validation_error}
	\end{subfigure}
	
	\caption{Comparison of simple PINN and LS-GD PINN results for a 30 Hz scattered wavefield simulation on Marmousi model. 
		(a) The selected portion of the Marmousi velocity model. 
		(b) The FD solution. 
		(c) The simple PINN prediction. 
		(d) The LS-GD PINN prediction. 
		(e) Evolution of training losses. 
		(f) Evolution of validation errors, excluding the PML region.}
	\label{fig:marmousi_results}
\end{figure}

The second test evaluates the proposed method using a subset of the benchmark Marmousi velocity model (Figure~\ref{fig:marmousi_results}a). The Marmousi model \citep{versteeg1994marmousi} is a realistic and widely used synthetic velocity model in seismic imaging, characterized by strong lateral velocity variations and geological complexity, making it a challenging benchmark for wavefield simulation and inversion methods \citep{yang2025,wu2023}.

We consider a portion of the Marmousi model and simulate a 30 Hz scattered wavefield using FD modeling. The  FD solution is computed on a high-resolution grid to serve as the reference for PINN comparisons. To mitigate boundary reflections, we employ a large perfectly matched layer (PML) with $a_0=0.8$, which absorbs outgoing waves (Figure~\ref{fig:marmousi_results}b).

For this test, we design a deep neural network (DNN) with five hidden layers of 160 neurons each, incorporating an encoder layer with $K=5$. The network is trained using 67,500 randomly varying collocation points per epoch. The learning rate follows the same exponentially decaying schedule as in the previous test.

We use the PDE with PML (equation \ref{eq:pde_pml}) and compare two optimization methods: the simple PINN, which employs gradient descent (GD), and the proposed LS-GD PINN, which incorporates the LS solver during training. Both methods use the same network architecture, collocation points, and training parameters. The loss function for the simple PINN is defined using Equations~\ref{eq:loss_total} and \ref{eq:loss_pde_pml}, while the LS-GD PINN uses Equations~\ref{eq:LS_loss} and \ref{eq:DR_PML}.

Figure~\ref{fig:marmousi_results} presents the results of this experiment. The simple PINN starts to converge only after 30,000 epochs but exhibits instabilities, with frequent jumps in loss values. In contrast, the LS-GD method enables stable and rapid convergence from the beginning of training with smaller fluctuations in loss values (note the logarithmic scale of the vertical axis). After 150,000 epochs, the standard PINN still does not achieve the accuracy level reached by the LS-GD PINN in just 15,000 epochs. This demonstrates the significant efficiency improvement brought by the LS solver in PINN training.

\section{Discussion}

Physics-informed neural networks (PINNs) with a linear output layer rely on nonlinear feature learning in earlier hidden layers, while the last layer’s weights can be efficiently determined via least squares (LS). This reduces the complexity of the gradient descent (GD) optimization problem and accelerates convergence as our expriments show. Traditional PINNs trained solely with GD exhibit a low-frequency bias. This is partly due to the smoother gradients of low-frequency functions, making them easier to learn. Unlike GD, the LS solver directly estimates the optimal weights without favoring low frequency components. However, if the penultimate layer features lack high-frequency information, the LS solution inherits this limitation, even though the solver itself is unbiased.

Gradient descent (GD) updates network weights iteratively using randomly sampled batches of collocation points. Over many iterations, these stochastic updates aggregate information from different regions of the domain, enabling the network to approximate the solution globally. In contrast, the least-squares (LS) solver computes the output-layer weights analytically at each training step based solely on the current batch of collocation points. This can increase sensitivity to the spatial representativeness of each batch. Despite this, our results show that the proposed LS-enhanced training framework achieves high validation accuracy even when relatively few collocation points are used per epoch.

The computational cost of the LS solver remains low due to its operation on a small \( P \times P \)  matrix, where \( P \) is the number of neurons in the penultimate layer. While each training step involves \( N + N_C\) collocation points, typically on the order of thousands, the LS step is computationally efficient. Constructing the matrix \( \mathbf{D}^T \mathbf{D} \) requires \( \mathcal{O}((N + N_C) P^2) \) operations, while solving for the weights via Cholesky decomposition has a complexity of \( \mathcal{O}(P^3) \). Given that \( P \) is relatively small (on the order of tens), this additional cost is minimal compared to standard GD, making the LS-GD approach a practical enhancement that improves convergence speed without significant computational overhead.

Incorporating perfectly matched layers (PML) modifies the matrix representation used in the LS formulation, nearly doubling the size of the matrix \( \mathbf{D} \) (as defined in Equation~\ref{eq:DR_PML}) and increasing the size of the normal matrix to \( 2P \times 2P \). This raises the cost of constructing \( \mathbf{D}^T \mathbf{D} \) to \( \mathcal{O}(4 (2N + N_C) P^2) \), and the Cholesky decomposition cost to \( \mathcal{O}(8 P^3) \). Furthermore, since the physical domain expands due to the inclusion of a PML, the number of collocation points \( N \) must be increased to maintain wavefield accuracy, further impacting computational cost. Nevertheless, the LS step remains computationally feasible due to the moderate values of \( P \) typically used in practice.

\section{Conclusion}

For the Physics-Informed Neural Networks (PINNs) solution of the scattered Helmholtz equation, we introduced a hybrid least-squares and gradient descent (LS-GD) optimization framework. Our derived matrix-form formulation, which involves a new definition of the GD loss function, combined with key implementation strategies enable an efficient approach to scattered wavefield simulation, with or without the inclusion of Perfectly Matched Layers (PML). The LS solver analytically determines the optimal weights of the linear part of the network at each training step, accelerating convergence and mitigating the slow optimization typically associated with high-frequency wavefields.

Through numerical experiments on benchmark velocity models, we demonstrated that the LS-GD PINN achieves more stable and accurate solutions in fewer training iterations. In particular, our results show that the LS-enhanced training method allows for rapid convergence even with a limited number of collocation points. The method effectively handles complex wavefields, including a high-frequency wavefield in the challenging Marmousi model, where standard GD training struggles to converge.

The computational cost introduced by the LS solver remains minimal due to the small size of the least-squares system, making the approach practical for large-scale applications. Although the introduction of PML increases computational cost, our experiments indicate that the LS-GD framework remains efficient and scalable.

These findings suggest that integrating the LS solver into PINN training significantly enhances efficiency and convergence stability. Future work could explore adaptive strategies for selecting penultimate layer size based on problem-specific characteristics, as well as extensions to more complex wave propagation scenarios and three-dimensional models.

\section*{Acknowledgments}
This research was supported by th following Research Projects/Grants: European Union’s Horizon Europe research and innovation programme under the Marie Sklodowska-Curie grant agreement No 101119556.TED2021-132783B-I00 funded by MICIU/AEI /10.13039/501100011033 and by FEDER, EU; PID2023-146678OB-I00 funded by MICIU/AEI /10.13039/ 501100011033 and by the European Union Next Generation EU/ PRTR; “BCAM Severo Ochoa” accreditation of excellence CEX2021-001142-S funded by MICIU/AEI/ 10.13039/ 501100011033; Basque Government through the BERC 2022-2025 program; BEREZ-IA (KK-2023/00012) and RUL-ET(KK-2024/00086), funded by the Basque Government through ELKARTEK; Consolidated Research Group MATHMODE (IT1456-22) given by the Department of Education of the Basque Government; BCAM-IKUR-UPV/EHU, funded by the Basque Government IKUR Strategy and by the European Union Next Generation EU/PRTR. 

\section*{Code Availability}
A TensorFlow implementation of the proposed method is available at: \\
\url{https://github.com/mahdiabedi/least-squares-embedded-optimization-for-scattered-wavefield-simulation}

	\bibliographystyle{apalike}
	\bibliography{references}
	
	\end{document}